\documentclass[lettersize,journal]{IEEEtran}
\usepackage{amsmath,amsfonts}
\usepackage{algorithmic}
\usepackage{algorithm}
\usepackage{array}
\usepackage[caption=false,font=normalsize,labelfont=sf,textfont=sf]{subfig}
\usepackage{textcomp}
\usepackage{stfloats}
\usepackage{url}
\usepackage{verbatim}
\usepackage{graphicx}
\usepackage{cite}
\usepackage{multirow}
\usepackage{hyperref}
\usepackage{url}
\usepackage{tcolorbox}
\usepackage[utf8]{inputenc} 
\usepackage[T1]{fontenc}    
\usepackage{booktabs}       
\usepackage{amsfonts}       
\usepackage{nicefrac}       
\usepackage{microtype}      
\usepackage{xcolor}         
\usepackage{caption}
\usepackage{diagbox}
\usepackage{newfloat}
\usepackage{graphicx}
\usepackage{listings}
\usepackage{booktabs} 
\usepackage{graphicx}
\usepackage{amsmath}
\usepackage{amssymb}
\usepackage{booktabs}
\usepackage{multirow}
\usepackage{overpic}
\usepackage{amsmath}
\usepackage{overpic}
\usepackage{bm}
\usepackage{multicol}
\usepackage[switch]{lineno}
\usepackage{algorithm}
\usepackage{algorithmic}
\usepackage{bbding}
\usepackage{tabularx}
\usepackage{graphicx}
\usepackage{subcaption}
\usepackage{tabu}                     
\usepackage{multirow}                 
\usepackage{multicol}                 
\usepackage{multirow}                
\usepackage{float}                    
\usepackage{makecell}                 
\usepackage{booktabs} 
\usepackage{bbm}
\usepackage{bbding}
\usepackage{listings}
\usepackage{graphicx}
\usepackage{algorithm}
\usepackage{algorithmic}
\usepackage{tabularx}
\usepackage{setspace}
\usepackage{textcomp}
\usepackage{stfloats}
\usepackage{url}
\usepackage{verbatim}
\usepackage{graphicx}
\usepackage{cite}
\usepackage{microtype}      
\usepackage{color}
\usepackage{pifont}
\usepackage{subcaption}
\usepackage{hyperref}
\hypersetup{
    colorlinks=true,
    linkcolor=blue,
    filecolor=blue,      
    urlcolor=blue,
    citecolor=cyan,
}
\newcommand{\eg}{\textit{e}.\textit{g}.}

\newcommand{\ie}{\textit{i}.\textit{e}.}

\begin{document}


\title{Rethinking `Knowledge' in Distillation: An In-context Sample Retrieval Perspective}


\author{Jinjing Zhu,~\IEEEmembership{Student Member,~IEEE}, Songze Li,~\IEEEmembership{Member,~IEEE}, Lin Wang$^\dagger$,~\IEEEmembership{Member,~IEEE}
\thanks{J. Zhu is with the Artificial Intelligence Thrust, The Hong Kong University of Science and Technology (HKUST), Guangzhou, China. E-mail: \{jinjingzhu.mail@gmail.com\}}
\thanks{S. Li is with the School of Cyber Science and Engineering, Southeast University,  Nanjing, China. E-mail: \{songzeli@seu.edu.cn\}}
\thanks{L. Wang is with the School of Electrical and Electronic Engineering, Nanyang Technological University, Singapore. E-mail: linwang@ntu.edu.sg.}
\thanks{$^\dagger$Corresponding author}}

\markboth{Journal of \LaTeX\ Class Files,~Vol.~14, No.~8, August~2021}%
{Shell \MakeLowercase{\textit{et al.}}: A Sample Article Using IEEEtran.cls for IEEE Journals}


\maketitle

\begin{abstract}
Conventional knowledge distillation (KD) approaches are designed for the student model to predict similar output as the teacher model for \emph{each sample}. Unfortunately, the relationship across samples with same class--a crucial knowledge typically useful for 
KD -- is often neglected. In this paper, we explore to redefine the `knowledge' in distillation, capturing the relationship between each sample and its corresponding \emph{in-context samples} (\ie, a group of similar samples with the same or different classes), and perform KD from an \textit{in-context sample retrieval} perspective. {As KD is a type of learned label smoothing regularization (LSR), we first conduct a theoretical analysis showing that the teacher's knowledge from the in-context samples is a crucial contributor to regularize the student training with the corresponding samples.} 
Buttressed by the analysis, we propose a novel in-context knowledge distillation (\textbf{IC-KD}) framework that shows its superiority across diverse KD paradigms (offline, online, and teacher-free KD). 
Firstly, we construct a feature memory bank from the teacher model and retrieve in-context samples for each corresponding sample through retrieval-based learning. We then introduce Positive In-Context Distillation (\textbf{PICD}) to reduce the discrepancy between a sample from the student and the aggregated in-context samples with the same class from the teacher in the logit space. Moreover, Negative In-Context Distillation (\textbf{NICD}) is introduced to separate a sample from the student and the in-context samples with different classes from the teacher in the logit space. Extensive experiments demonstrate that IC-KD is effective across various types of KD, and consistently achieves state-of-the-art performance on CIFAR-100 and ImageNet datasets.
\end{abstract}


\IEEEpubidadjcol

\section{Introduction}
Over the past few decades, deep neural networks (DNNs) have achieved remarkable success in various computer vision tasks, such as image classification~\cite{simonyan2014very, he2016deep, zhu2023patch}, object detection~\cite{zou2023object, zhao2019object}, and semantic segmentation~\cite{long2015fully, chen2018encoder}. However, DNNs are often coupled with increasing computational costs, which limit their application to resource-limited devices. Therefore, many efforts have been made to compress model size without sacrificing much accuracy~\cite{mehta2021mobilevit, wu2022tinyvit}.
Knowledge Distillation (KD)~\cite{WangY22,SonNCH21, ParkCJKH21} -- one of the model compression techniques -- aims to transfer knowledge from a powerful but cumbersome teacher model into a compact yet effective student model.
KD has demonstrated significant success in various tasks, such as image classification~\cite{liu2023norm, hinton2015distilling} and semantic segmentation~\cite{zhu2023good, yang2022cross}.
The key intuition of KD lies in how to effectively capture and transfer the knowledge of a pre-trained teacher model for training a student model.

\begin{figure}[t]
\captionsetup{font=small}
\begin{center}
\includegraphics[width=0.99\linewidth]{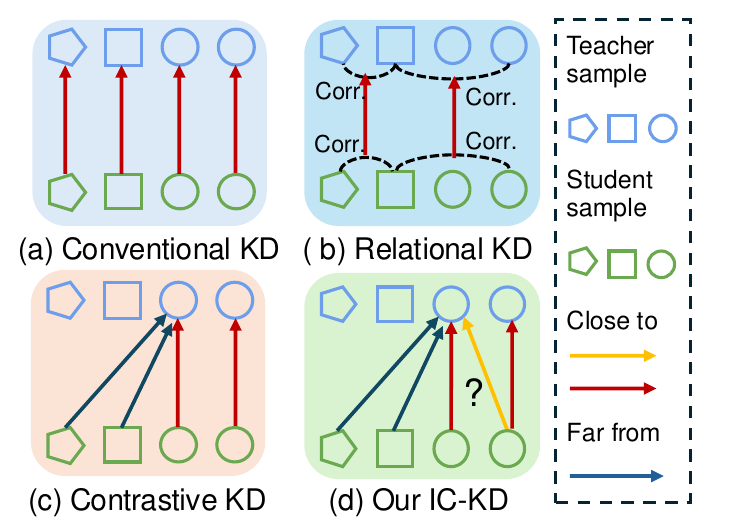}
\end{center}
\caption{
Illustration of four typical approaches: classic KD, Relation-based KD, Contrastive KD, and our IC-KD. Note that different shapes represent samples of different classes.}
\label{cover_fig}
\end{figure}

The mainstream KD paradigms can be categorized into: 1) logits-based~\cite{hinton2015distilling, liu2023norm, zhang2024goodsam}, 2) feature-based~\cite{zagoruyko2016paying, shu2021channel, liu2021exploring}, and 3) relation-based~\cite{peng2019correlation, yang2022cross, tian2019contrastive, tung2019similarity, huang2022knowledge}. 
Among the relation-based methods (as show in Fig.~\ref{cover_fig}b), CIRKD~\cite{yang2022cross}, SP~\cite{tung2019similarity}, CC~\cite{peng2019correlation}, and RKD~\cite{park2019relational} explore the knowledge of correlations among instances or pixels for knowledge transfer. 
Moreover, CRD~\cite{tian2019contrastive} and CRCD~\cite{zhu2021complementary} propose contrastive distillation that only captures the knowledge of correlations between different class probabilities, as illustrated in Fig.~\ref{cover_fig}c. 
Nonetheless, a crucial knowledge -- \textit{the relationship across samples with same class} -- might be ignored but typically beneficial for KD (Fig.~\ref{cover_fig}d).

\begin{figure}
    \centering
    \includegraphics[width=0.89\linewidth]{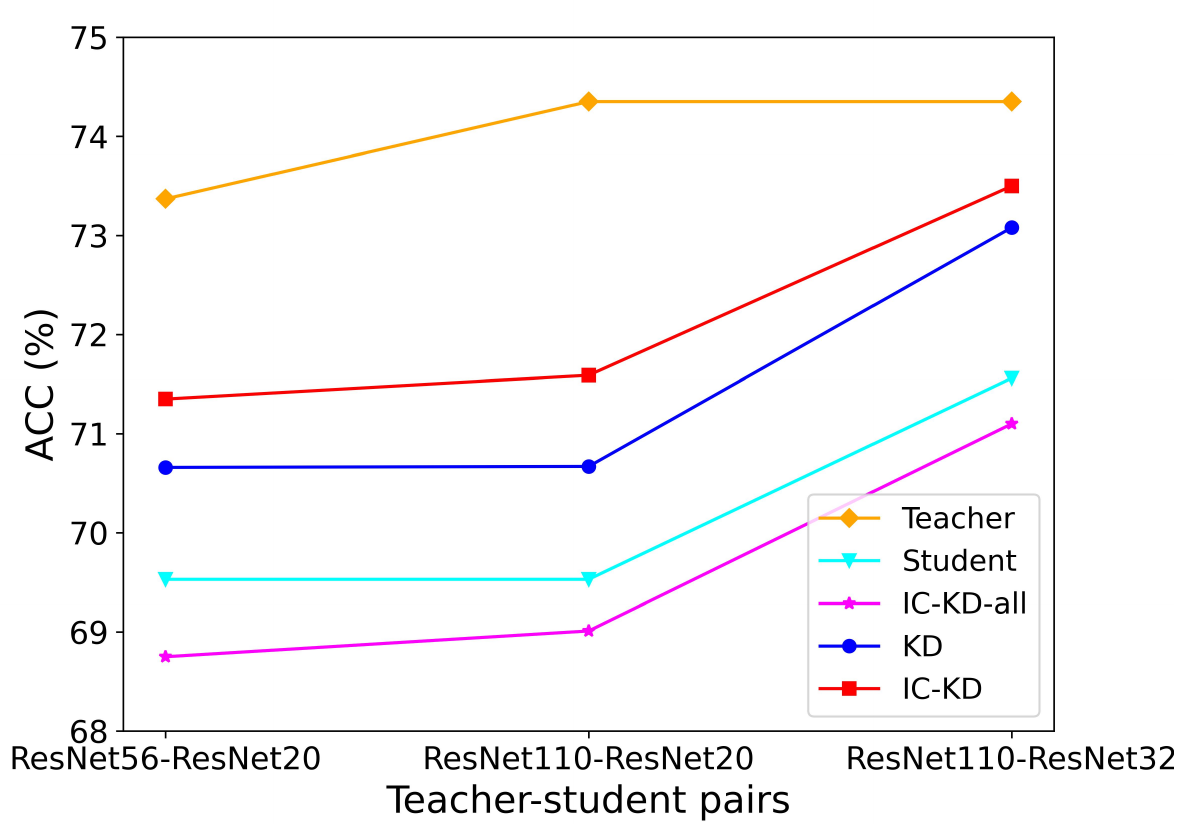}
     \caption{Comparative results of various teacher-student pairs on CIFAR-100. Note that IC-KD-all refers to the method of knowledge distillation by capturing the relationships across all samples of the same class within the logit space.} 
    \label{second_fugure}
\end{figure}

To validate such an assumption, we conduct a series of experiments, as depicted in Fig.~\ref{second_fugure}.
Our findings indicate that the student's performance decreases when these relationships across all samples with same class are considered. The substantial discrepancy in logit differences among all samples with same class in the two models appears to negatively impact the effectiveness of knowledge transfer.

\textcolor{black}{This further motivates us to redefine `knowledge' transferred across samples within the same or different classes in distillation. 
Drawing on the experimental results and analysis above, we pose a question: given a query sample, is it possible to retrieve optimal similar samples and then leverage the correlation across these paired samples to facilitate better knowledge transfer? To probe such samples, we define the group of similar samples with the same or different classes as \emph{in-context samples} for the given query sample, and utilize these in-context samples to capture both inter- and intra-class correlations.}  

To answer the question, we first revisit knowledge distillation via label smoothing regularization (LSR) and theoretically analyze the rationale for regularizing the student training using in-context samples. Prior arts~\cite{muller2019does, yuan2020revisiting, chandrasegaran2022revisiting} investigate the compatibility or relationship between LSR and KD. Notably, Tf-KD~\cite{yuan2020revisiting} considers KD as a form of learned LSR as it delivers a regularization effect in training a student model. Given that LSR can enhance model performance, and KD serves as a learned form of LSR by training the student model with soft targets from the teacher model, what if we employ additional soft targets to regularize the student training? This raises two novel technical challenges: 1) how to retrieve in-context samples for each sample? and 2) how to effectively regularize student training using these in-context samples.

To address these challenges, our key idea is to leverage the teacher model to retrieve optimal in-context samples for regularizing the model's training with LSR. To this end, we propose a novel in-context knowledge distillation (\textbf{IC-KD}) to regularize the student training with in-context samples. Firstly, we construct a feature memory bank from the fixed teacher model and employ the retrieval-based learning approach to search the in-context samples for each sample. With the in-context samples, we then propose two regularization loss terms: Positive In-Context Distillation (\textbf{PICD}) and Negative In-Context Distillation (\textbf{NICD}). Specifically, PICD aims to regularize the student training by decreasing the discrepancy between one sample from the student and the aggregated corresponding in-context samples with the same class from the teacher model in the logit space. To address the over-fitting issue easily caused by solely relying on the constraint from positive in-context samples, NICD enhances the student learning by pushing apart the student model's samples from the corresponding in-context samples of different classes from the teacher in logit space. 

Our IC-KD exhibits generality capacity and can be readily applied to train the student model across various KD settings, including different KD variants, backbones, and tasks. We validate the effectiveness of the proposed approach through extensive experiments on CIFAR-100 and ImageNet datasets. In summary, the main contributions of our paper are as follows:
\begin{itemize}
    \item We rethink `knowledge' in distillation via label smoothing regularization from an in-context sample retrieval perspective and then introduce a novel IC-KD framework.  
    \item To facilitate knowledge transfer, we propose two key components of in-context KD: positive in-context distillation (PICD) and negative in-context distillation (NICD).
    \item Our IC-KD is highly versatile, demonstrating the effectiveness to be generalized across various variants, tasks, and datasets. 
\end{itemize}

\section{Related Work}

\textbf{Knowledge distillation} (KD) has emerged as a method to transfer knowledge from a large teacher model to a compact student model via minimizing the discrepancy between their representations~\cite{gou2021knowledge, wang2021knowledge, hinton2015distilling, zhu2024source}. KD has achieved remarkable success in various tasks, such as image classification~\cite{liu2023norm, zhao2022decoupled}, semantic segmentation~\cite{liu2019structured, wang2020intra, zhu2023good}, and object detection~\cite{wang2023crosskd}. 
Based on how the knowledge is defined, current KD methods can also be divided into  1) logits-based, 2) feature-based, and 3) relation-based approaches. For a more detailed review of KD, we refer readers to recent surveys, \eg,~\cite{wang2021knowledge,liu2021data}.
Logits-based methods use the output logits from a teacher model as soft supervision for guiding a student model~\cite{hinton2015distilling, liu2023norm}.
Feature-based KD focuses on capturing appropriate transferred knowledge, such as the intermediate feature maps~\cite{adriana2015fitnets, zhu2023good, chen2022knowledge} or their transformations~\cite{zagoruyko2016paying, shu2021channel, liu2021exploring}. Relation-based KD leverages the relations between data samples or pixels in the feature space for distillation~\cite{peng2019correlation, yang2022cross, tian2019contrastive, tung2019similarity, park2019relational}. In particular, RKD~\cite{peng2019correlation}, SP~\cite{tung2019similarity}, and CC~\cite{peng2019correlation} capture the knowledge of relationships between anchor instances or pixels with others for knowledge transfer. CIRKD~\cite{yang2022cross} introduces cross-image relational KD, utilizing semantic relations among pixels for transfer. CRD~\cite{tian2019contrastive} introduces contrastive learning into distillation and transfers relationships between samples with different classes. However, these relation-based KD approaches are designed for the student model to predict a similar output as the teacher model for \emph{each sample}. The relationship across samples within the same class--a crucial knowledge typically useful for KD -- is often neglected.
 \textit{Therefore, in this work, we redefine `knowledge' in distillation via in-context sample retrieval approach, aiming to capture the relationship between each sample and its corresponding \emph{in-context samples}, and perform KD via label smoothing regularization}. 

\textbf{Retrieval-based learning} has garnered significant attention in both the natural language processing (NLP) and computer vision (CV) communities. In NLP, efforts are directed towards enhancing the performance of LLM~\cite{liu2023reta, zhang2023retrieve} and text generation~\cite{cheng2024lift, cai2022recent}. In CV, researchers exploit retrieval-based learning to address real-world challenges such as image retrieval tasks~\cite{datta2008image, deselaers2008features, saito2023pic2word}. Additionally, there has been substantial interest in integrating cross-modal alignment and retrieval tasks, exemplified by endeavors in image-text retrieval~\cite{fu2023learning, qu2021dynamic} and text-video retrieval~\cite{croitoru2021teachtext}. More recently, retrieval-based training has been employed to conduct nearest neighbor retrieval from annotated features prompts for in-context scene understanding~\cite{balazevic2024towards}, and to automate the selection of optimal in-context samples in visual in-context learning~\cite{zhang2024makes}. \textit{Motivated by the success of leveraging in-context samples or features to enhance task performance without fine-tuning models, our approach  builds upon pioneering work in learning to distill knowledge by acquiring in-context samples for each sample through retrieval-based learning and subsequently regularizing the student training with the relationship between each sample and its in-context samples.}

\section{Method}
\label{method}





\begin{figure*}[t]
\captionsetup{font=small}
\begin{center}
\includegraphics[width=0.9\linewidth]{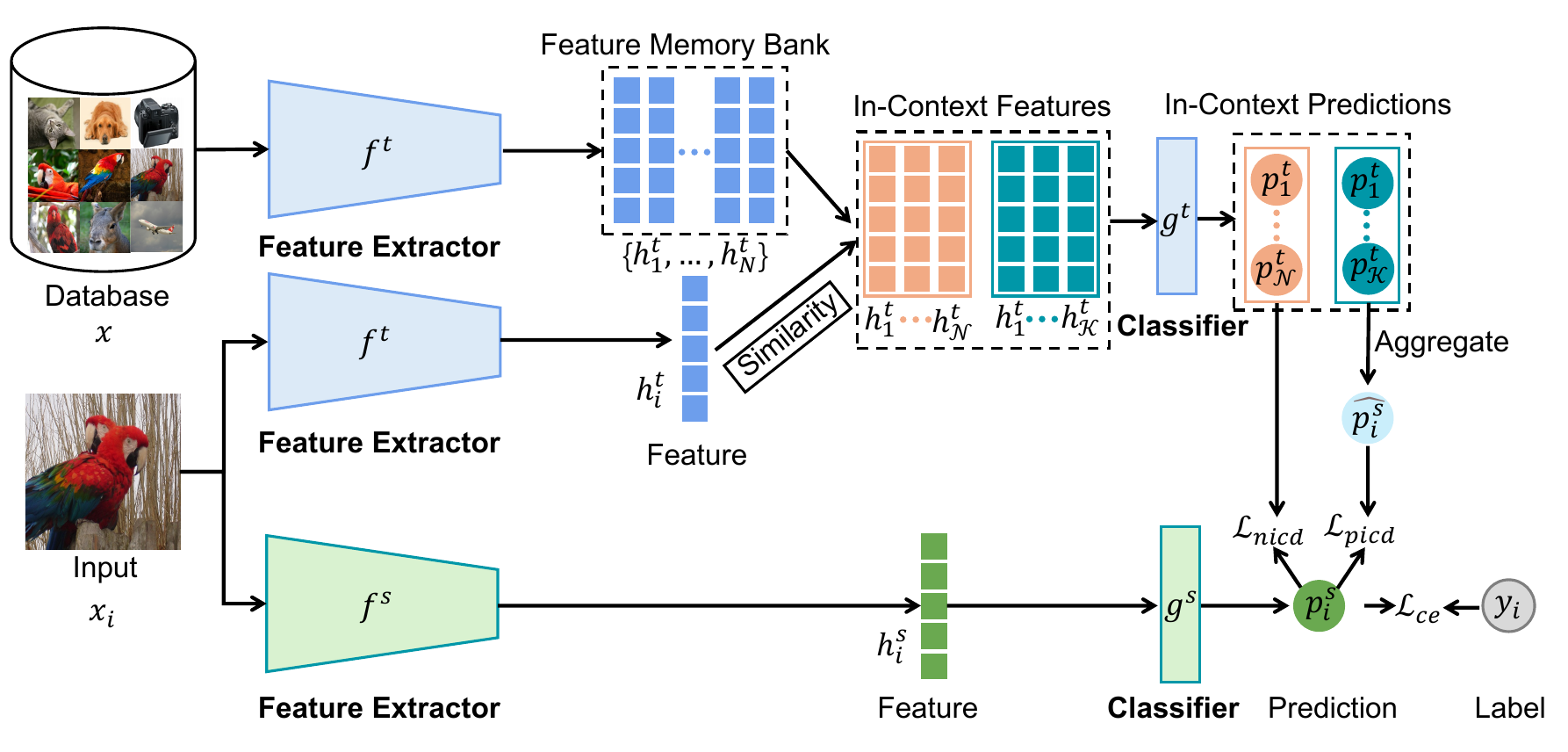}
\end{center}

\caption{Overview of our framework. It consists of two feature extractors ($f^t$ and $f^{s}$) and two classifiers ($g^s$ and $g^t$). Given a sample $x_i$, we extract its feature $h_{i}^s$, and obtain predictions $p^s_{i}$ from the corresponding classifier $g^s$. Moreover, we build a feature memory bank \{$\boldsymbol{h}_{1}^t$, ..., $\boldsymbol{h}_{j}^t$\} using all the database with the feature extractor $f^t$, and obtain the feature $h^t_i$ of sample $x_i$. Then we calculate the similarity of feature $h^t_i$ and feature memory bank, and utilize the ground truth of samples to identify the Top-$\mathcal{K}$ and Top-$\mathcal{N}$ similar features (in-context features) from same and different classes, respectively. Finally, we regularize the student training with two proposed loss terms: $\mathcal{L}_{picd}$ and $\mathcal{L}_{nicd}$.}

\label{framework}
\end{figure*}

\subsection{Rethinking `Knowledge' in Distillation}
\label{rethinking}
Given a training dataset $(X, Y)$ comprising training samples $x_i$ ($i \in\{1 \ldots N\}$) and their corresponding labels $y_i$, we denote the feature extractors of the student and teacher as $f^s$ and $f^t$, respectively. The features $\boldsymbol{h}_{i}^s = f^s(x_i)$ and $\boldsymbol{h}_{i}^t = f^t(x_i)$ are extracted from the sample $x_i$ using the respective feature extractor. The feature $\boldsymbol{h}_{i}^s$ is subsequently input into the classifier $\boldsymbol{g}^s$ to obtain the logits $\boldsymbol{z}_{i}^s$ =  $\boldsymbol{g}^s(\boldsymbol{h}_{i}^s)$. Similarly, logits $\boldsymbol{z}_{i}^t$ are obtained by the teacher's classifier $\boldsymbol{g}^t$ and are formulated as $\boldsymbol{g}^t(\boldsymbol{h}_{i}^t)$. Finally, $\boldsymbol{p}_{i}^s =  softmax (\boldsymbol{z}_{i}^s)$ and $\boldsymbol{p}_{i}^t =  softmax(\boldsymbol{z}_{i}^t)$ represent the final prediction probabilities of student and teacher, respectively. Note that $\boldsymbol{p}_{i}^t(k, \tau) = softmax\left(\boldsymbol{z}_{i}^t(k)\right)=\frac{\exp \left(\boldsymbol{z}_{i}^t(k)/ \tau\right)}{\sum_{m=1}^K \exp \left(\boldsymbol{z}_{i}^t (m)/ \tau\right)}$, where $k \in\{1 \ldots K\}$ is the class of data and $\tau$ is the temperature to soften $\boldsymbol{p}_{i}^t$. $\boldsymbol{p}_{i}^t(k, \tau)$ denotes the out prediction of the sample $x_i$ with the teacher model.

Prior studies~\cite{muller2019does, yuan2020revisiting, chandrasegaran2022revisiting} explore the compatibility or relationship between LSR and KD. Notably, Tf-KD~\cite{yuan2020revisiting} interprets KD as a variant of learned LSR, suggesting that KD imposes a regularization influence during student model training. Given the ground truth distribution $y(k)$, the modified label distribution $q(k)$ in LSR is defined as
\begin{equation}
\label{lsr1}
    \boldsymbol{q}(k)=(1-\alpha) y(k)+\alpha u(k), 
\end{equation}

where $u(k)$ represents a uniform distribution as 1/K, and $\alpha$ is a trade-off parameter. Consequently, the objective of LSR is reformulated as
\begin{equation}
\small
    \mathcal{L}_{lsr}=-\sum_{k=1}^K \boldsymbol{q}(k) \log \boldsymbol{p}(k)=(1-\alpha) H(y, \boldsymbol{p})+\alpha H(u, \boldsymbol{p}),
\end{equation}
where $H(\cdot)$ is cross-entropy loss. And the objective of KD is defined as
\begin{equation}
\label{objective of KD}
\begin{aligned}
    \mathcal{L}_{kd}&=(1-\alpha) H(y, \boldsymbol{p}^s)+\alpha K L\left(\boldsymbol{p}^t, \boldsymbol{p}^s\right) \\
    &=(1-\alpha) H(y, \boldsymbol{p}^s)+\alpha H\left(\boldsymbol{p}^t, \boldsymbol{p}^s\right).
\end{aligned}
\end{equation}

Note that $KL\left(\boldsymbol{p}^t, \boldsymbol{p}^s\right)=H\left(\boldsymbol{p}^t, \boldsymbol{p}^s\right)-H\left(\textbf{p}^t\right)$, $H\left(\textbf{p}^t\right)$ remains constant and does not influence the training of the student model. By setting the temperature $\tau=1$, we obtain $\mathcal{L}_{kd}=H\left(\hat{\textbf{q}}^t, \textbf{p}\right)$, where $\hat{\boldsymbol{q}}^t$ is
\begin{equation}
\label{kd21}
    \hat{\boldsymbol{q}}^t(k)=(1-\alpha) \boldsymbol{q}(k)+\alpha \boldsymbol{p}^t(k).
\end{equation}


By comparing the objectives of LSR (Eq. \ref{lsr1}) and KD (Eq. \ref{kd21}), we observe that KD manifests as a particular instance of learned LSR, aimed at regularizing the learning of the student model by leveraging the predictions from the teacher model. This suggests that the `knowledge' distilled in this process essentially serves as a form of `regularization' for student training. However, existing KD methods regularize student training using the same sample, neglecting constraints that arise from different samples within the same or even different classes. Consequently, an essential inquiry arises: \textit{can we utilize the teacher's predictions derived from other samples and leverage them to enact effective regularization for student training with each input sample?}

Considering the experimental results and analysis in Fig.~\ref{second_fugure}, and to answer the above question, we rethink the regularization term to obtain similar teacher's predictions from other samples for effectively training the student model with each input sample. Our approach involves obtaining similar samples, termed in-context samples, for each individual sample utilizing a feature memory bank extracted from the teacher. Note that the discrepancy between the teacher’s predictions for in-context samples and the student’s predictions for corresponding samples decreases as the similarity between these paired samples increases. Therefore, by leveraging both similarity between feature $h_{i}^s$ and feature memory bank, and ground truth, we identify $\mathcal{K}$ positive in-context samples ($x_1, ..., x_j$, $j \in \mathcal{K}$) within the same class for the sample $x_i$. Subsequently, we aggregate the logit predictions from these positive in-context samples to derive the aggregated predictions $\hat{\boldsymbol{p}}_i^s$ (as defined in Eq.~\ref{aggregate}). Finally, the training objective for the student model with sample $x_i$ is formulated accordingly.

\begin{equation}
    \mathcal{L}_{kd}(x_i)=(1-\alpha) H(y_i, \boldsymbol{p}_i^s)+\alpha H\left(\hat{\boldsymbol{p}}_i^s, \boldsymbol{p}_i^s\right). 
\end{equation}

Moreover, to mitigate the over-fitting issue easily caused by solely relying on the constraint from positive in-context samples, by following the retrieval of $\mathcal{N}$ negative in-context samples ($x_1, ..., x_j$, $j \in \mathcal{N}$) from different classes for sample $x_i$, we improve the discrepancy between these paired samples as
\begin{equation}
\small
    \mathcal{L}_{kd}(x_i)=(1-\alpha) H(y_i, \boldsymbol{p}_i^s)-\alpha \frac{1}{\mathcal{N}} \sum_{j \sim \mathcal{N}} H\left(\boldsymbol{p}_j^t, \boldsymbol{p}_i^s\right). 
\end{equation}

Building upon the aforementioned objectives involving in-context samples, this work aims to ascertain the optimal regularization strategy for training the student model by leveraging retrieval-based learning to obtain these in-context samples.

\subsection{Distillation via In-Context Sample Retrieval }
\label{in-context sample retrieval}
To retrieve optimal in-context samples and effectively utilize them for student training regularization alongside each corresponding sample, we propose In-Context Knowledge Distillation (\textbf{IC-KD}) via employing in-context sample retrieval.





\noindent \textbf{Positive In-Context Distillation.} 
We introduce Positive In-Context Distillation (\textbf{PICD}) as a regularization technique for enhancing the student training, incorporating positive in-context samples from the same class. To retrieve these positive in-context samples for each sample, we employ the teacher's feature extractor $f^t$ to form a feature memory bank \{$\boldsymbol{h}_{1}^t$, ..., $\boldsymbol{h}_{j}^t$\}, where $j \in\{1 \ldots N\}$ represents all training data. Given a sample $x_i$, its feature representation $\boldsymbol{h}_{i}^t$ serves as a query to cross-attend over the feature memory bank with temperature ${\beta}_{1}$. The similarity between the sample $x_i$ and other samples $x_j$ is calculated as $s_{i, j} = \frac{1}{{\beta}_1} \frac{\left\langle\boldsymbol{h}^t_{i}, \boldsymbol{h}_{j}^t\right\rangle}{\left\|\boldsymbol{h}^t_{i}\right\|\left\|\boldsymbol{h}_j^t\right\|}$.  Incorporating the label information, matrix $\mathcal{M} (x_i, x_j)$ indicates whether $x_i$ and $x_j$ share the same label. Top-$\mathcal{K}$ positive in-context samples for the sample $x_i$ are identified using Rank $\mathcal{R}_1^{\mathcal{K}}$. After obtaining these positive in-context samples for the sample $x_i$, we aggregate their predictions ${\boldsymbol{p}}_j^t$, weighted by the similarity $a_{i,j}$. 
Thus, the process of obtaining the positive in-context samples and aggregated in-context predictions $\hat{\boldsymbol{p}}_i^s$ for sample $x_i$ is defined as
{\setlength\abovedisplayskip{2pt}
\setlength\belowdisplayskip{4pt}
\begin{equation}
\label{aggregate}
\small
\begin{aligned}
      & s_{i, j}=\frac{1}{{\beta}_1} \frac{\left\langle\boldsymbol{h}^t_{i}, \boldsymbol{h}_{j}^t\right\rangle}{\left\|\boldsymbol{h}^t_{i}\right\|\left\|\boldsymbol{h}_j^t\right\|}, \quad 
\hat{s_{i,j}} = {\mathcal{R}_1^{\mathcal{K}}}(s_{i,j}* \mathcal{M}_{i,j}), \\
&a_i=\operatorname{softmax}\left(\hat{s_i}\right),\quad \hat{\boldsymbol{p}}_i^s=\sum_j a_{i, j}\boldsymbol{p}_{j}^t({\tau}_{1}),
\end{aligned}
\end{equation}}

where ${\tau}_{1}$ is a temperature used to soften ${\boldsymbol{p}}_j^t$ and $\textbf{p}_i^s$ for effective regularization. Based on these aggregated in-context predictions, we propose regularizing the student training with positive in-context samples with KL divergence as

\begin{equation}
  \mathcal{L}_{picd} = KL (\hat{\boldsymbol{p}}_i^s,\boldsymbol{p}_i^s (\tau_{1})). 
\end{equation}
\noindent \textbf{Negative In-Context Distillation.}
In addition to leveraging positive in-context samples for student training regularization, we further incorporate negative in-context samples to regularize the student training, thereby preventing over-fitting. Specifically, we introduce negative in-context distillation (\textbf{NICD}) to regularize the student training by separating a sample from the student and the corresponding in-context samples with different classes from the teacher in the logit space. We now describe details below. 

For a given sample $x_i$, we calculate the similarity $d_{i, j}$ between its feature $\boldsymbol{h}_{i}^t$ and the feature memory bank using a temperature parameter $\beta_{2}$. Leveraging the similarity $d_{i, j}$ and the matrix $\mathcal{M} (x_i, x_j)$, we identify the corresponding negative in-context samples for each sample $x_i$ with a designated Rank $\mathcal{R}_2$. To enhance the generalization ability in distinguishing samples across different classes and based on experimental results, we employ Rank $\mathcal{R}_2$ to select samples with distinct classes within one epoch as negative in-context samples. Further details regarding these definitions are



{\setlength\abovedisplayskip{1pt}
\setlength\belowdisplayskip{1pt}
\begin{equation}
\begin{aligned}
\small
& d_{i, j}=\frac{1}{{\beta}_2} \frac{\left\langle\boldsymbol{h}^t_{i}, \boldsymbol{h}_{j}^t\right\rangle}{\left\|\boldsymbol{h}^t_{i}\right\|\left\|\boldsymbol{h}_j^t\right\|}, \quad 
\hat{d_{i,j}} = {\mathcal{R}_2}(d_{i,j}* (1-\mathcal{M}_{i,j})),\\
&b_i=\operatorname{softmax}\left(\hat{d_i}\right).
\end{aligned}
\end{equation}}


After retrieving the negative in-context samples \{$x_j$, j = 1,..., $\mathcal{N}$\} for the sample $x_i$, we aim to increase the discrepancy between the predictions of the sample and its corresponding negative in-context samples, thereby enhancing student training regularization. Drawing from empirical insights provided by CRD~\cite{tian2019contrastive} and infoNCE~\cite{oord2018representation}, we employ cosine similarity rather than cross-entropy to guide the regularization process with negative in-context samples. And the regularization term is formulated as
\begin{equation}
   \mathcal{L}_{nicd}=1-\cos \left(\boldsymbol{p}_i^s, \boldsymbol{p}_i^t\right)+ b_{i,j}\cos \left(\boldsymbol{p}_i^s, \boldsymbol{p}_j^t\right).
\end{equation}

\subsection{Total objective}
Building on the aforementioned components, the objective of our proposed IC-KD encompasses the task loss, positive in-context distillation loss, and negative in-context distillation loss as follows:
\begin{equation}
   \mathcal{L}= \mathcal{L}_{ce} +\mathcal{L}_{kd}+ {\gamma}_{picd}*\mathcal{L}_{picd}+{\gamma}_{nicd}*\mathcal{L}_{nicd}.  
\end{equation}

Here, $\mathcal{L}_{ce}$ denotes the task training loss with sample $x_i$, while ${\gamma}_{picd}$ and ${\gamma}_{nicd}$ are trade-off parameters used to balance the task training loss and the regularization loss terms. The pseudo-algorithm of the proposed method for offline KD is shown in Algorithm~\ref{alg1}.

\begin{algorithm}[]
	\caption{The proposed framework for offline KD} 
	\label{alg1} 
	\begin{algorithmic}[1]
	    \STATE \textbf{Input}: $\{X,Y\}$; max iterations: $T$
	    \\ Teacher model: $f(\theta^t)$;
             \\ Student model: $f(\theta^{s})$;
	    \STATE  \textbf{Initialization}: Set $\theta^s$;\\
            \STATE Obtain the feature memory bank \{$\boldsymbol{h}_{1}^t$, ..., $\boldsymbol{h}_{j}^t$\} using the teacher model $f(\theta^t)$;
            \STATE Obtain the Top-$\mathcal{K}$ positive in-context samples for sample $x_i$ with Rank $\mathcal{R}_1^{\mathcal{K}}$:\\
            $s_{i, j}=\frac{1}{{\beta}_1} \frac{\left\langle\boldsymbol{h}^t_{i}, \boldsymbol{h}_{j}^t\right\rangle}{\left\|\boldsymbol{h}^t_{i}\right\|\left\|\boldsymbol{h}_j^t\right\|}, \quad 
            \hat{s_{i,j}} = {\mathcal{R}_1^{\mathcal{K}}}(s_{i,j}* \mathcal{M}_{i,j}) $.\\
            \STATE Obtain the negative in-context samples for sample $x_i$ with Rank $\mathcal{R}_2$:\\
            $d_{i, j}=\frac{1}{{\beta}_2} \frac{\left\langle\boldsymbol{h}^t_{i}, \boldsymbol{h}_{j}^t\right\rangle}{\left\|\boldsymbol{h}^t_{i}\right\|\left\|\boldsymbol{h}_j^t\right\|}, \quad 
            \hat{d_{i,j}} = {\mathcal{R}_2}(d_{i,j}* (1-\mathcal{M}_{i,j}))$.\\
	    \FOR{t $\xleftarrow[]{}$ 1 to $T$}
    	    \STATE Attain the prediction $\boldsymbol{p}_i^s$ for input samples;
                \STATE Calculate the loss of positive in-context distillation with the aggregated in-context predictions $\hat{\boldsymbol{p}}_i^s$:\\
               $ a_i=\operatorname{softmax}\left(\hat{s_i}\right)$,\\
               $\hat{\boldsymbol{p}}_i^s=\sum_j a_{i, j}\boldsymbol{p}_{j}^t({\tau}_{1}),$\\
               $\mathcal{L}_{picd} = KL (\hat{\boldsymbol{p}}_i^s,\boldsymbol{p}_i^s (\tau_{1}))$.\\
               \STATE Calculate the loss of negative in-context distillation with the samples with distinct classes within one epoch:\\
               $b_i=\operatorname{softmax}\left(\hat{d_i}\right)$,\\
               $\mathcal{L}_{nicd}=1-\cos \left(\boldsymbol{p}_i^s, \boldsymbol{p}_i^t\right)+b_{i,j}\cos \left(\boldsymbol{p}_i^s, \boldsymbol{p}_j^t\right).$

               \STATE Calculate the loss of total objective:\\
              $ \mathcal{L}= \mathcal{L}_{ce} +\mathcal{L}_{kd}+ {\gamma}_{picd}*\mathcal{L}_{picd}+{\gamma}_{nicd}*\mathcal{L}_{nicd}.  $
               
         \STATE Back propagation for $\mathcal{L}$;
         \STATE Update the student $\theta^s$.
    	 \ENDFOR
	    \STATE  \textbf{return}  $\theta^s$.
	    \STATE  \textbf{End}.
	\end{algorithmic} 
\end{algorithm}

\section{Experiments}
\label{experiments}
To evaluate the effectiveness of our proposed IC-KD, we conduct extensive experiments across three distinct knowledge distillation variants: 1) offline KD, 2) online KD, and 3) teacher-free KD.

\begin{table*}[t]
  \caption{Comparison of Top-1 mean accuracy (\%) on CIFAR-100 using same type network architectures for teacher and student models. IC-KD+CRD represents combinations of IC-KD with CRD~\cite{tian2019contrastive}. Best and second-best results are highlighted in bold and underlined, respectively. We use $\Delta$ to show performance gain over CRD.}
  \centering
    \resizebox{0.99\linewidth}{!}{
  \begin{tabular}{l|l|lllllll}
    \toprule
    \multirow{4}{*}{Distillation mechanism}&
    Teacher & WRN-40-2 &WRN-40-2 &ResNet56 &ResNet110& ResNet110& ResNet32x4&VGG13\\
    &Student & WRN-16-2& WRN-40-1& ResNet20 &ResNet20& ResNet32 &ResNet8x4 &VGG8\\
    \cmidrule{2-9} 
    &Teacher & 76.31 &76.31& 73.37& 74.35& 74.35& 79.56& 75.07\\
    &Student &73.80& 71.70& 69.53 &69.53 &71.56 &72.87 &70.75\\
    \midrule
    \multirow{3}{*}{Logits-based}&
    KD~\cite{hinton2015distilling} &74.92 &73.54 &70.66 &70.67& 73.08& 73.33 &72.98\\
    &DKD~\cite{zhao2022decoupled}&\textbf{76.24}&\textbf{74.81}&\textbf{71.97}&n/a&\underline{74.11}&76.32&74.68\\
   & KD+LSKD~\cite{sun2024logit}&76.11&74.37&71.43&71.48&\textbf{74.17}&76.62&74.36\\
    \midrule
    \multirow{8}{*}{Feature-based}&
    NST~\cite{huang2017like} &73.68 &72.24 &69.60 &69.53 &71.96 &73.30& 71.53\\
    &FT~\cite{kim2018paraphrasing} &73.25& 71.59& 69.84 &70.22& 72.37 &72.86 &70.58\\
    &PKT~\cite{passalis2018learning} &74.54 &73.45 &70.34 &70.25 &72.61 &73.64& 72.88\\
    &AB~\cite{heo2019knowledge} &72.50 &72.38 &69.47& 69.53& 70.98 &73.17 &70.94\\
   &ReviewKD~\cite{chen2021distilling}& 76.12&75.09&71.89&71.34&73.89&75.63&74.84\\
    &SimKD~\cite{chen2022knowledge}&75.53&74.53&71.05& 71.06&73.92&\textbf{78.08}&\underline{74.89}\\
    & NORM~\cite{liu2023norm}&75.57 & 74.72& 70.37& 70.85&73.43& 76.64&73.88 \\
   & CAT-KD~\cite{guo2023class}&75.60&\textbf{74.82}&71.62&n/a&73.62&76.91&74.65\\
    \midrule
    \multirow{2}{*}{Relation-based}&
    SP~\cite{tung2019similarity} &73.83 &72.43 &69.67 &70.04 &72.69 &72.94 &72.68\\
   & CC~\cite{peng2019correlation} &73.56& 72.21& 69.63& 69.48& 71.48& 72.97& 70.71\\
    \midrule
    \multirow{5}{*}{Contrastive learning-based}&
    CRD~\cite{tian2019contrastive} &75.48 &74.14 &71.16& 71.46 &73.48& 75.51& 73.94\\
   & IC-KD&\underline{76.21}& 74.49& 71.35& \underline{71.59} &73.50 &\underline{77.22} &74.81\\
   &$ \Delta$&\textcolor{red}{+0.73}&\textcolor{red}{+0.35}&\textcolor{red}{+0.19}&\textcolor{red}{+0.13}&\textcolor{red}{+0.02}&\textcolor{red}{+1.71}&\textcolor{red}{+0.87}\\
  \cmidrule{2-9} 
    &IC-KD+CRD&76.07&74.19&\underline{71.67}&\textbf{71.75} &73.92&76.76&\textbf{74.90}\\
    &$ \Delta$&\textcolor{red}{+0.59}&\textcolor{red}{+0.05}&\textcolor{red}{+0.51}&\textcolor{red}{+0.29}&\textcolor{red}{+0.44}&\textcolor{red}{+1.25}&\textcolor{red}{+0.96}\\
    \bottomrule
  \end{tabular}}
     \label{cifar-100-same}
    
\end{table*}

\begin{table*}[t]
  \caption{Top-1 mean accuracy (\%) comparison on CIFAR-100 employing distinct type network architectures for teacher and student models. We use $\Delta$ to show performance gain over CRD.}
  \centering
    \resizebox{0.99\linewidth}{!}{
  \begin{tabular}{l|l|llllll}
    \toprule
    \multirow{4}{*}{Distillation mechanism}&
    Teacher & VGG13& ResNet50 &ResNet50& ResNet32x4& ResNet32x4& WRN-40-2\\
    &Student & MobileNetV2& MobileNetV2 &VGG8& ShuffleNetV1 &ShuffleNetV2&ShuffleNetV1\\
    \cmidrule{2-8} 
    &Teacher&74.64& 79.23& 79.23& 79.42& 79.42& 75.61\\
   & Student&64.81 &64.81 &70.75 &71.63& 72.96 &71.63\\
    \midrule
     \multirow{3}{*}{Logits-based}&
    KD~\cite{hinton2015distilling} &67.37& 67.35& 73.81 &74.07 &74.45 &74.83\\
    &DKD~\cite{zhao2022decoupled}&69.71&70.35&n/a&76.45&77.07&76.70\\
     &KD+LSKD~\cite{sun2024logit}&68.61& 69.02&n/a&n/a&75.56&n/a\\
    \midrule
    \multirow{8}{*}{Feature-based}&
    NST~\cite{huang2017like} &58.16& 64.96& 71.28 &74.12& 74.68 &74.89\\
    &FT~\cite{kim2018paraphrasing} &61.78& 60.99& 70.29& 71.75 &72.50 &72.03\\
    &PKT~\cite{passalis2018learning}& 67.13& 66.52 &73.01 &74.10& 74.69 &73.89\\
    &AB~\cite{heo2019knowledge}  &66.06& 67.20& 70.65 &73.55& 74.31& 73.34\\
   & ReviewKD~\cite{chen2021distilling}&\textbf{70.37}&69.89&n/a&n/a&77.78&n/a\\
    &SimKD~\cite{chen2022knowledge}&69.44&69.97&n/a&n/a&\underline{78.39}&n/a\\
  &  NORM~\cite{liu2023norm}&68.94& \underline{70.56} &\underline{75.17}& \underline{77.42}&78.07& 77.06\\
     & CAT-KD~\cite{guo2023class}&69.13&\textbf{71.36}&n/a&\textbf{78.26}&\textbf{78.41}&\underline{77.35}\\

    \midrule

    \multirow{2}{*}{Relation-based}&
      SP~\cite{tung2019similarity} &66.30& 68.08 &73.34& 73.48& 74.56& 74.52\\
   & CC~\cite{peng2019correlation}  &64.86& 65.43 &70.25& 71.14 &71.29& 71.38\\
    \midrule 
    \multirow{5}{*}{Contrastive learning-based}&
    CRD~\cite{tian2019contrastive} &69.73& 69.11& 74.30& 75.11& 75.65 &76.05\\
    &IC-KD&\underline{70.10}&69.78&75.15&76.74&77.52&\textbf{77.42}\\
    &$\Delta$&\textcolor{red}{+0.37}&\textcolor{red}{+0.67}&\textcolor{red}{+0.85}&\textcolor{red}{+1.63}&\textcolor{red}{+1.87}&\textcolor{red}{+1.37}\\
    \cmidrule{2-8} 
    &IC-KD+CRD&69.29&70.28&\textbf{75.29}&76.90&77.45&77.03\\
   & $\Delta$&\textcolor{green}{-0.44}&\textcolor{red}{+1.17}&\textcolor{red}{+0.99}&\textcolor{red}{+1.79}&\textcolor{red}{+1.80}&\textcolor{red}{+0.98}\\
    \bottomrule
  \end{tabular}}

  \label{cifar-100-different}

\end{table*}

\begin{table*}[t]
\caption{Top-1 accuracy (\%) comparison on ImageNet.}
  \centering
  \resizebox{0.99\linewidth}{!}{
  \begin{tabular}{l|l|lllllllllllllll}
    \toprule
    Teacher & Student&KD &AT &OFD& RKD &CRD &SRRL &SemCKD &ReviewKD &SimKD &DistPro &DKD&NORM&LSKD+KD&IC-KD\\
    \midrule
    ResNet34 (73.31) &ResNet18 (70.13)  &70.68& 70.59 &71.08 &71.34& 71.17 &71.73& 70.87 &71.61 &71.66& 71.89&71.70& \underline{72.14}&71.42&\textbf{72.36}\\
    ResNet50 (76.13) &MobileNet (69.63) &70.68 &70.72 &71.25 &71.32 &71.40& 72.49 &n/a &72.56& n/a &73.26& n/a&\underline{74.26}&n/a&\textbf{74.58}\\
    \bottomrule
  \end{tabular}}
      \label{imagenet}
\end{table*}

\subsection{Settings}
\textbf{Datasets.} We conduct experiments using two widely recognized datasets: CIFAR-100~\cite{krizhevsky2009learning} and ImageNet~\cite{deng2009imagenet}. CIFAR- 100~\cite{krizhevsky2009learning} comprises 50,000 training images and 10,000 test images across 100 classes. And ImageNet~\cite{deng2009imagenet} provides 1.2 million training images from 1,000 classes, with an additional 50,000 images for validation.

\textbf{Implementations.} For offline KD, using CRD settings~\cite{tian2019contrastive} with 240 training epochs, we conduct experiments on 7 or 6 teacher-student pairs on CIFAR-100, employing either the same or different architectural styles. Each experiment is repeated three times, and we report the top-1 mean recognition rate on the test set. To ensure fair comparisons, we adhere to the same training settings as CRD~\cite{tian2019contrastive} for our method. Specifically, each teacher-student pair is trained using the stochastic gradient descent (SGD) optimizer for 240 epochs, with a batch size of 64, a weight decay of 0.0005, and a momentum of 0.9. All models are trained on  a single NVIDIA GeForce RTX 3090 GPU.

On the ImageNet dataset, we follow the experimental setup outlined by CRD ~\cite{tian2019contrastive}, employing two widely used teacher-student pairs. During training, input images are first resized to 256 × 256 pixels, followed by random cropping to 224 × 224 pixels or their horizontal flips. The models are trained using the stochastic gradient descent (SGD) optimizer for 100 epochs, with a batch size of 256, a weight decay of 0.0001, and a momentum of 0.9. The initial learning rate is set to 0.1. All experiments are conducted on four NVIDIA GeForce RTX 3090 GPUs.


For teacher-free KD, we adhere to the experimental framework established in Tf-KD~\cite{yuan2020revisiting}, conducting seven tasks across different network architectures, including GoogLeNet~\cite{szegedy2015going} and DenseNet~\cite{huang2017densely}. The baseline models are trained for 200 epochs with a batch size of 128. The initial learning rate is set to 0.1 and is subsequently reduced by a factor of 5 at epochs 60, 120, and 160. The training process employs the stochastic gradient descent (SGD) optimizer, with a momentum of 0.9 and a weight decay of 5e-4. In the teacher-free KD setting, \textit{note that the trained baseline model serves as the teacher, while the untrained model functions as the student.} Additional details regarding this setup are provided in the seminal work on Tf-KD~\cite{yuan2020revisiting}.

For online KD, following the settings of KDCL~\cite{guo2020online}, we conduct seven student-student pair tasks. All models are trained for 200 epochs, starting with an initial learning rate of 0.1, which is reduced by a factor of 10 at epochs 100 and 150. The weight decay is set to 0.0005, the batch size to 128, and the momentum to 0.9. During training, all images are padded with 4 pixels, and a 32 × 32 crop is randomly sampled from the padded images or their horizontal flips. \textit{Note that in the online KD setting, as the two students are trained iteratively, the feature memory bank, derived from student 1, evolves with each epoch of training}.

For a comprehensive comparison, we evaluate our method against a broad range of current mainstream offline KD techniques. These include vanilla KD~\cite{hinton2015distilling}, DKD~\cite{zhao2022decoupled}, LSKD~\cite{sun2024logit}, NST~\cite{huang2017like}, FT~\cite{kim2018paraphrasing}, PKT~\cite{passalis2018learning}, AB~\cite{heo2019knowledge}, ReviewKD~\cite{chen2021distilling}, SimKD~\cite{chen2022knowledge}, NORM~\cite{liu2023norm}, CAT-KD~\cite{guo2023class}, SP~\cite{tung2019similarity}, CC~\cite{peng2019correlation}, CRD~\cite{tian2019contrastive}, AT~\cite{zagoruyko2016paying}, OFD~\cite{heo2019comprehensive}, RKD~\cite{passalis2018learning}, SRRL~\cite{yang2021knowledge}, SemCKD~\cite{chen2021cross}, and DistPro~\cite{deng2022distpro}. 

\subsection{Results and Discussion}

\textbf{Results on CIFAR-100.} Tab.~\ref{cifar-100-same} presents a comparative analysis of our IC-KD approach against existing KD methods on the CIFAR-100 dataset, involving 7 teacher-student pairs with identical network architectures. Similarly, Tab.~\ref{cifar-100-different} provides the results on 6 teacher-student pairs with different type network architectures. Notably, IC-KD consistently outperforms the contrastive learning-based CRD~\cite{tian2019contrastive} across all pairs, achieving over \textbf{1}\% improvement in some tasks. Additionally, integrating IC-KD with CRD as IC-KD+CRD further enhances performance across almost all tasks, demonstrating seamless integration with recent KD techniques. Moreover, IC-KD shows comparable or superior performance to logits-based, relation-based, and feature-based methods, effectively highlighting the benefits of leveraging inter- and intra-class in-context samples for the student training regularization.

\textbf{Results on ImageNet.} 
To comprehensively assess the effectiveness of IC-KD in enhancing student training through an in-context sample retrieval, we conduct experiments on the ImageNet dataset. The results, presented in Tab. \ref{imagenet}, demonstrate the superiority of IC-KD over previous KD methodologies. Specifically, in the ResNet34 $\to$ ResNet18 teacher-student setting, IC-KD improves the student's accuracy from 70.13\% to 72.36\%, exceeding NORM by \textbf{0.24}\%. Furthermore, in the ResNet50 $\to$ MobileNet pair, IC-KD outperforms NORM~\cite{liu2023norm} by \textbf{0.32}\%, achieving an impressive \textbf{74.58}\% accuracy with MobileNet. These findings underscore the substantial enhancement in student performance facilitated by our proposed regularization losses incorporating in-context samples.

\textbf{Ablation studies of IC-KD.} In Tab. \ref{ablation study}, an ablation study is conducted to assess the contributions of distinct components within IC-KD with same type network architectures. The key findings are as follows: (1) The PICD loss ($\mathcal{L}_{picd}$) effectively regulates student training, leading to performance enhancements across all seven teacher-student pairs. Particularly noteworthy is the ResNet32x4 $\to$ ResNet8x4 pair, where the student model achieves a significant accuracy boost of 2.28\% accuracy compared to using only the KD loss. (2) The NICD loss ($\mathcal{L}_{nicd}$) consistently improves performance across all pairs, with the ResNet32x4 $\to$ ResNet8x4 pair showing a notable accuracy gain of 3.77\% accuracy. (3) Combining both regularization loss terms generally amplifies student performance, highlighting the effectiveness of incorporating in-context samples into student training regularization. (4) However, in the ResNet110 $\to$ ResNet32 pair, employing the combined terms yields inferior performance compared to using solely $\mathcal{L}_{picd}$, underscoring the importance of balancing these terms to optimize student performance. Furthermore, we conduct an ablation study to examine the contribution of each component of IC-KD across different network architectures, as presented in Tab.~\ref{ablation-loss-different}.

\textbf{Impact of weights $a_{i}$ and $b_i$.} In the design $\gamma_{picd}$ and $\gamma_{nicd}$, we leverage the similarity between sample features and their corresponding in-context samples to weight the regularization term. To evaluate the impact of these weights, $a_i$ and $b_i$, we conduct experiments as detailed in Tab.~\ref{ablation study-a-b}. Specifically, when using the weights $a_{i}$ and $b_i$, our method outperforms the baseline on 6 tasks and 5 tasks, respectively, compared to the scenario without these weights. The results consistently demonstrate that the inclusion of the weights generally enhances performance across most pairs, underscoring the positive impact of weighting the regularization terms.


\begin{table*}[t]
    \caption{Ablation studies on the proposed loss terms on CIFAR-100 employing same type network architectures for teacher
and student models.} 
    \label{ablation study}
  \centering
  \resizebox{0.99\linewidth}{!}{
  \begin{tabular}{lll|llllllll}
    \toprule
    \multirow{4}{*}{$\mathcal{L}_{kd}$} &
    \multirow{4}{*}{$\mathcal{L}_{picd}$} &
    \multirow{4}{*}{$\mathcal{L}_{nicd}$} &
    Teacher & WRN-40-2 &WRN-40-2 &ResNet56 &ResNet110& ResNet110& ResNet32x4&VGG13\\
    &&&Student & WRN-16-2& WRN-40-1& ResNet20 &ResNet20& ResNet32 &ResNet8x4 &VGG8\\
    \cmidrule{4-11} 
    &&&Teacher & 76.31 &76.31& 73.37& 74.35& 74.35& 79.56& 75.07\\
    &&&Student &73.80& 71.70 &69.53& 69.53& 71.56 &72.87 &70.75\\
    \midrule
    \checkmark&&&Student&74.92 &73.54 &70.66 &70.67& 73.08& 73.33 &72.98\\
\checkmark&\checkmark&&Student&75.85&74.10&71.32&71.00&\textbf{73.77}&75.61&73.35\\
     \checkmark&&\checkmark&Student&76.10&74.30&70.16&70.82&73.25&77.10&74.67\\
    \checkmark&\checkmark&\checkmark&Student&\textbf{76.21}&\textbf{74.49}&\textbf{71.35}&\textbf{71.59} &73.50&\textbf{77.22}&\textbf{74.81}\\
    \bottomrule
  \end{tabular}}
\end{table*} 

\begin{table*}[t]
  \centering
   \caption{Ablation studies on the proposed loss terms on CIFAR-100 employing distinct type network architectures for teacher and student models.}
   \label{ablation-loss-different}
  \resizebox{0.99\linewidth}{!}{
  \begin{tabular}{lll|lllllll}
    \toprule
    \multirow{4}{*}{$\mathcal{L}_{kd}$} &
    \multirow{4}{*}{$\mathcal{L}_{picd}$} &
    \multirow{4}{*}{$\mathcal{L}_{nicd}$} &
    Teacher & VGG13& ResNet50 &ResNet50& ResNet32x4& ResNet32x4& WRN-40-2\\
    &&&Student &MobileNetV2& MobileNetV2 &VGG8& ShuffleNetV1 &ShuffleNetV2&ShuffleNetV1\\
    \cmidrule{4-10} 
    &&&Teacher & 75.07& 79.23& 79.23& 79.42& 79.42& 76.31\\
    &&&Student &64.81 &64.81 &70.75 &71.63& 72.96 &71.63\\
    \midrule
    \checkmark&&&Student& 67.37& 67.35 &73.81& 74.07 &74.45& 74.83 \\
\checkmark&\checkmark&&Student&68.13&69.63&74.66&74.02&75.10&75.06\\
     \checkmark&&\checkmark&Student&70.01&69.40&74.76&76.09&77.73&76.73\\
    \checkmark&\checkmark&\checkmark&Student&70.10&69.78&75.15&76.74&77.52&77.36\\
    \bottomrule
  \end{tabular}}
\end{table*}

\begin{table*}[t]
    \caption{Impact of the weights $a_{i}$ and $b_{i}$ in $\mathcal{L}_{picd}$ and $\mathcal{L}_{nicd}$ on CIFAR-100. The omission of $a_i (b_i)$ implies that weighting is not required for calculating $\mathcal{L}_{picd}$ ($\mathcal{L}_{nicd}$).} 
    \label{ablation study-a-b}
  \centering
  \small
  \resizebox{0.99\linewidth}{!}{
  \begin{tabular}{l|llllllll}
    \toprule
    &Teacher & WRN-40-2 &WRN-40-2 &ResNet56 &ResNet110& ResNet110& ResNet32x4&VGG13\\
    &Student & WRN-16-2& WRN-40-1& ResNet20 &ResNet20& ResNet32 &ResNet8x4 &VGG8\\
    \midrule
   $\mathcal{L}_{picd}$ (w/ $a_i$)&Student& 
   \textbf{75.85}&\textbf{74.10}&\textbf{71.32}&71.00&\textbf{73.77}&\textbf{75.61}&\textbf{73.35}\\
   $\mathcal{L}_{picd}$ (w/o $a_i$)&Student&74.93&73.41&70.91&\textbf{71.37}&73.45&74.95&72.47\\
    \midrule
    $\mathcal{L}_{nicd}$ (w/ $b_i$)&Student&\textbf{76.10}&74.30&\textbf{70.16}&\textbf{70.82}&73.25&\textbf{77.10}&\textbf{74.67}\\
    $\mathcal{L}_{nicd}$ (w/o $b_i$)&Student&75.97&\textbf{74.88}&70.08&70.65&\textbf{73.48}&76.92&74.08\\
    \bottomrule
  \end{tabular}}
    
\end{table*}

\begin{figure*}[t]
    \centering
    \includegraphics[width=0.95\linewidth]{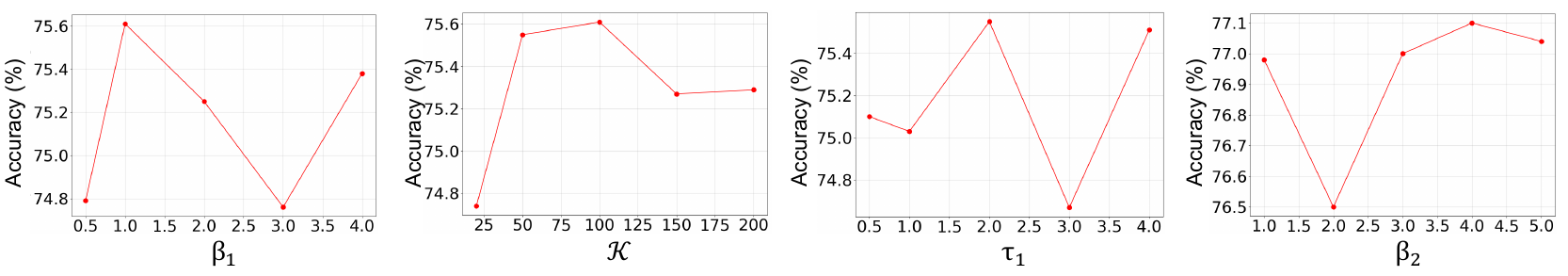}

    \caption{Effects of varying value of $\beta_{1}$, $\mathcal{K}$, $\tau_{1}$, and $\beta_{2}$ on CIFAR-100 with ResNet32x4 $\to$ ResNet8x4.}
      \label{figure: ablation}
\end{figure*}

\begin{figure}[t]
    \centering
    \includegraphics[width=0.99\linewidth]{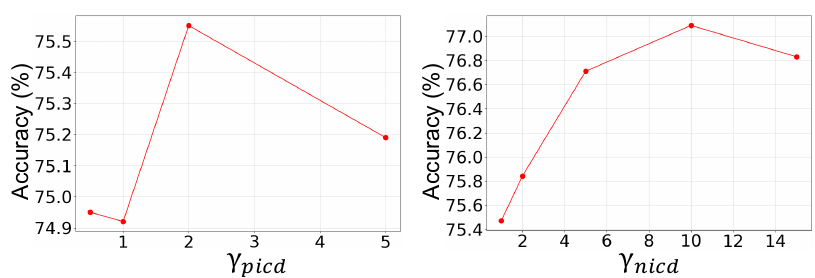}
 
    \caption{Effects of varying the value of ${\gamma}_{picd}$ and ${\gamma}_{nicd}$ on CIFAR-100 with ResNet32x4$\to$ResNet8x4.}
      \label{trade-off parameter}
  
\end{figure}

\begin{figure*}[t]
    \centering
    \includegraphics[width=0.99\linewidth]{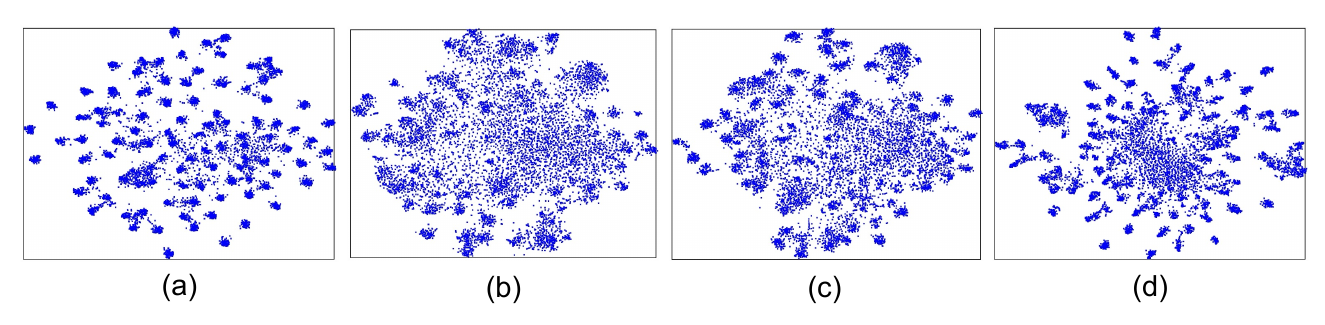}
    \caption{The visualization of features using different models on CIFAR-100. (a) ResNet32x4, (b) ResNet8x4, (c) ResNet8x4 trained with KD, and (d) ResNet8x4 trained with IC-KD. }
    \label{figure:tsne}
\end{figure*}

\begin{figure*}[t]
    \centering
    \includegraphics[width=0.99\linewidth]{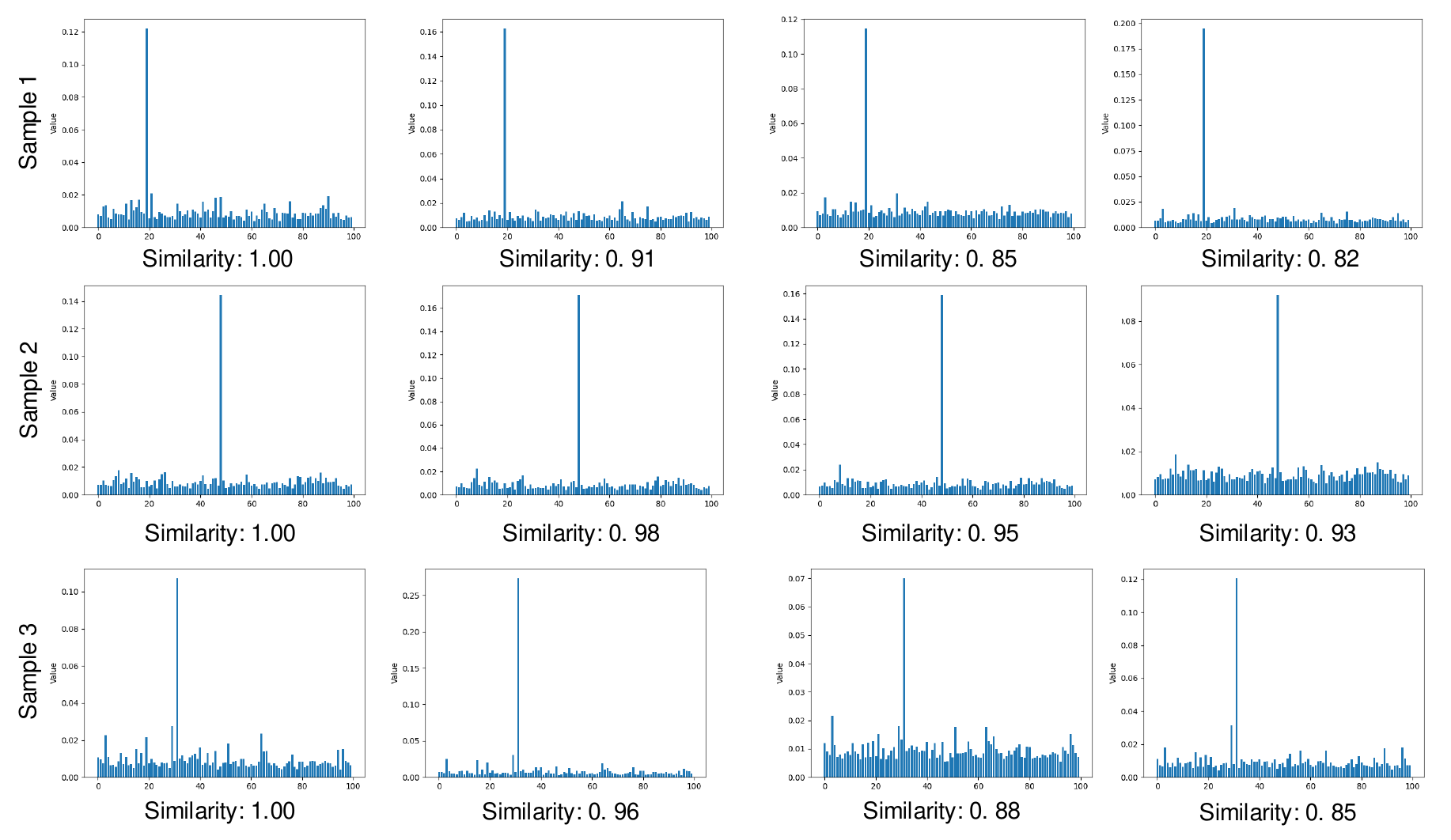}
    \caption{Visualization of logits on CIFAR-100. The first column displays logit predictions for samples, while subsequent columns present the logit predictions for corresponding positive in-context samples.}
    \label{figure:visual1}
\end{figure*}

\begin{table}[t]
    \caption{Training time (per epoch) vs. accuracy on CIFAR-100 with with ResNet32x4 $\to$ ResNet8x4.}
  \centering
  \setlength{\tabcolsep}{25pt}
    \resizebox{0.99\linewidth}{!}{
  \begin{tabular}{l|llll}
    \toprule
     Method & Accuracy&Time\\
    \midrule
    KD ~\cite{hinton2015distilling}&73.33&17.32s\\
    CRD~\cite{tian2019contrastive}& 75.51&23.34s\\
    Norm~\cite{liu2023norm}&76.64&17.82s\\
    \midrule
    IC-KD&77.22&20.83s\\
    \bottomrule
  \end{tabular}}

  \label{computer-cost}
\end{table}

\textbf{Sensitivity study for $\beta_{1}$, $\mathcal{K}$,  $\tau_{1}$, and $\beta_{2}$.} In designs of $\mathcal{L}_{picd}$ and $\mathcal{L}_{nicd}$, IC-KD incorporates four hyperparameters: $\beta_{1}$ and $\beta_{2}$, which soften the similarity and then weight in-context distillation loss terms, $\mathcal{K}$, which controls the number of optimal positive in-context samples, and $\tau_{1}$, which moderates the softening of the student and teacher predictions. As illustrated in Fig. \ref{figure: ablation}, the observations are as follows: (1) Increasing $\beta_{1}$ from 0.5 to 1 results in a 0.80\% accuracy gain, but further increases in $\beta_{1}$ degrade student performance. (2) The increase in $\mathcal{K}$ exhibits a peak in student performance at $\mathcal{K} = 100$, highlighting the importance of selecting optimal in-context samples for enhancing the student model and confirming the necessity of reliable teacher predictions for student regularization. (3) The variation in student performance across different values of temperature $\tau_{1}$ suggests that appropriate softening of predictions enhances student training across diverse samples and models. (4) Optimal performance for $\beta_{2}$, which weights the discrepancy between samples and their corresponding negative in-context samples, is observed at $\beta_{2}=4$, underscoring the significance of weighting the discrepancy for effective regularization with negative in-context samples.

\textbf{Parameter sensitivity of $\gamma_{picd}$ and $\gamma_{nicd}$.} 
Experiments in Fig.~\ref{trade-off parameter} are conducted to explore the impact of trade-off parameters $\gamma_{picd}$ and $\gamma_{nicd}$. The results highlight the critical role of both parameters in regulating student training and improving the student performance. Notably, optimal performance is observed when the student is trained with $\gamma_{picd}=2$ and $\gamma_{nicd}=10$. Based on comparisons across various values of trade-off parameters, these settings, $\gamma_{picd}=2$ and $\gamma_{nicd}=10$, are chosen as the default configurations.

\textbf{Visualization results.} Our IC-KD approach accomplishes the objective of KD by facilitating student training with in-context samples, leading to the convergence of the student's feature distribution towards that of the teacher. To illustrate this, we employ IC-KD using ResNet32x4 $\to$ ResNet8x4 on the CIFAR-100 validation set and visualize the feature distributions using t-SNE~\cite{van2008visualizing}. The comparative visualization results depicted in Fig.~\ref{figure:tsne} underscore the effectiveness of IC-KD in enabling the student to learn knowledge from the teacher.
 
\textbf{Training efficiency.} We evaluate the training time costs of various distillation methods in Tab.~\ref{computer-cost}. Our IC-KD approach strikes the most favorable balance between model performance and training time. Notably, IC-KD constructs a feature memory bank using only the teacher model before training the student model, thereby reducing the time required in comparison to CRD~\cite{tian2019contrastive}. 

\textbf{Visualization of positive in-context samples.} We present visualizations of positive in-context samples, along with their corresponding feature similarity and logit predictions, on CIFAR-100 in Fig.~\ref{figure:visual1}. From these visualizations of logits and cosine similarity, it is evident that the discrepancy between the teacher's predictions for in-context samples and the student's predictions for the corresponding samples diminishes as the similarity between the paired samples increases. This observation further reinforces the underlying rationale of our method, highlighting the effectiveness of knowledge distillation from the perspective of in-context sample retrieval.

\begin{table*}[t!]
  \caption{Comparison of Top-1 mean accuracy (\%) on CIFAR-100 for teacher-free KD.}
  \centering
    \resizebox{0.99\linewidth}{!}{
  \begin{tabular}{l|lllllll}
    \toprule
    Model & MobileNetV2&ShuffleNetV2&ResNet18&ResNet50&GoogLeNet&DenseNet121&ResNeXt29\\
    \midrule
    Baseline &68.36&70.11&75.75&79.11&78.45&78.64&81.15\\
    \midrule
    Tf-KD$_{self}$~\cite{yuan2020revisiting} &\underline{70.33}&71.55&\underline{77.42}&\underline{80.37}&\underline{80.53}&\underline{80.24}&\underline{82.16}\\
    Tf-KD$_{reg}$~\cite{yuan2020revisiting} &\textbf{70.36}&\underline{71.59}&76.55&78.99&78.92&78.75&81.84\\
    IC-KD&69.77 &\textbf{73.20}&\textbf{77.77}&\textbf{80.93} &\textbf{81.02}&\textbf{81.30}&\textbf{82.73}\\
    \bottomrule
  \end{tabular}}
     \label{teacher-free}
    
\end{table*}

\begin{table*}[t!]
   \caption{Comparison of Top-1 mean accuracy (\%) on CIFAR-100 for online KD.}
  \centering
    \resizebox{0.99\linewidth}{!}{
  \begin{tabular}{l|llllllll}
    \toprule
    \multirow{2}{*}{Model}& Student 1 & WRN-40-2 &WRN-40-2 &ResNet56 &ResNet110& ResNet110& ResNet32x4&VGG13\\
   &Student 2 & WRN-16-2& WRN-40-1& ResNet20 &ResNet20& ResNet32 &ResNet8x4 &VGG8\\
    \midrule
   \multirow{2}{*}{Model} &Student 1 & 76.31 &76.31& 73.37& 74.35& 74.35& 79.56& 75.07\\
    &Student 2 &73.80& 71.70 &69.53& 69.53& 71.56 &72.87 &70.75\\
    \midrule
    \multirow{2}{*}{DML~\cite{zhang2018deep}} &Student 1&\underline{78.28}&\underline{78.04}&74.14&\underline{76.24}&\underline{76.56}&\underline{80.31}&\underline{76.75}\\
    &Student 2&\underline{75.22}&73.47&\underline{70.27}&70.45&\underline{72.81}&\underline{74.74}&\underline{73.27}\\
    \midrule
    \multirow{2}{*}{KDCL~\cite{guo2020online}}&Student 1&78.14&77.80&\underline{74.27}&75.97&76.17&79.98&76.46\\
    &Student 2&74.91&\underline{73.79}&\underline{70.27}&\underline{70.57}&72.71&74.15&72.78\\
    \midrule
    \multirow{2}{*}{IC-KD}&Student 1&\textbf{78.84}&\textbf{78.05}&\textbf{75.03}&\textbf{76.62}&\textbf{76.79}&\textbf{80.40}&\textbf{77.01}\\
    &Student 2&\textbf{75.51}&\textbf{74.09}&\textbf{70.81}&\textbf{70.71}&\textbf{73.01}&\textbf{75.12}&\textbf{74.09}\\
    \bottomrule
  \end{tabular}}
    \label{online}
\end{table*}

\begin{table*}[t]
  \centering
    \caption{Performance comparison with state-of-the-art distillation methods on Cityscapes. The results of prior KD methods are reported in CIRKD~\cite{yang2022cross}.}
  \resizebox{0.99\linewidth}{!}{
  \begin{tabular}{l|lllllll}
    \toprule
    Teacher & Student&SKD~\cite{liu2019structured}&IFVD~\cite{wang2020intra}&CWD~\cite{shu2021channel}&CIRKD~\cite{yang2022cross}&IC-KD\\
    \midrule
   DeepLabV3-ResNet101 |78.07 &DeepLabV3-ResNet18 |74.21 & 75.42&75.59&75.55&\underline{76.38}&\textbf{76.89}\\
    DeepLabV3-ResNet101 |78.07 &PSPNet-ResNet18 |72.55&73.29&73.71&74.36&\underline{74.73}&\textbf{74.96} \\
    \bottomrule
  \end{tabular}}
  \label{segmentation}
\end{table*}

\begin{table}[t]
    \caption{Top-1 mean accuracy (\%) comparison on CIFAR-100 employing distinct heterogeneous network architectures for teacher and student models.}
  \centering
  \setlength{\tabcolsep}{25pt}
    \resizebox{0.99\linewidth}{!}{
  \begin{tabular}{l|ll}
    \toprule
    Teacher & Swin-L& Swin-L \\
    Student & ResNet50& Swin-T\\
    \midrule
    Teacher&86.3&86.3\\
    Student&78.5&81.3\\
    \midrule
    KD~\cite{hinton2015distilling}&80.0&81.5\\
    RKD~\cite{park2019relational}&78.9&81.2\\
    DIST~\cite{huang2022knowledge}&\underline{80.2}&\underline{82.3}\\
    \midrule
    IC-KD&\textbf{80.4}&\textbf{82.6}\\
    \bottomrule
  \end{tabular}}
  \label{cifar-100-vit}
\end{table}

\textbf{Knowledge distillation between heterogeneous models.} 
To further evaluate the efficacy of our IC-KD approach on more complex tasks, we conduct experiments using heterogeneous network architectures for the teacher and student models. As presented in Tab. \ref{cifar-100-vit}, we construct two teacher-student pairs using Swin-Transformer~\cite{liu2021swin}. The results demonstrate that our IC-KD method yields improvements, even with stronger models, underpinned by the in-context sample retrieval approach.

\textbf{Results on teacher-free KD.} 
In Tab. \ref{teacher-free}, our IC-KD is compared with two teacher-free knowledge distillation methods, Tf-KD$_{self}$ and Tf-KD$_{reg}$~\cite{yuan2020revisiting}. Tf-KD$_{self}$ lets the student to learn from a pre-trained student model, while Tf-KD$_{reg}$ enables the student to learn from a manually designed smooth distribution. From the comparative analysis, it is evident that our IC-KD methods outperform the alternatives on 6 out of 7 tasks, showcasing the effectiveness and versatility of our proposed IC-KD. Specifically, ShuffleNetV2 with IC-KD achieves an accuracy of 73.20\%, representing a 1.61\% improvement over Tf-KD$_{reg}$. 

\textbf{Results on online knowledge distillation.} To assess the generality of our proposed IC-KD, we conduct experiments on the online KD setting, as presented in Tab.~\ref{online}. Our comparison involves IC-KD against DML~\cite{zhang2018deep} and KDCL~\cite{guo2020online} methods, revealing that IC-KD consistently outperforms these prior methods across all student-student pairs. Specifically, for the ResNet56 $\to$ ResNet20 pair, our IC-KD achieves an accuracy of 75.03\% on ResNet56 and 70.81\% on ResNet20, underscoring the efficacy of IC-KD on online KD. Moreover, the results further demonstrate that retrieving in-context samples using a dynamic feature memory bank, extracted from the student during training, effectively regularize two students' training.

\textbf{Knowledge distillation on semantic segmentation.} To further assess the efficacy of our IC-KD approach on complex tasks, we conduct experiments using the Cityscapes dataset~\cite{cordts2016cityscapes}, which comprises 5000 finely annotated images, with 2975 images for training, 500 for validation, and 1525 for testing. The segmentation performance is evaluated across 19 classes. Our experiments involve two teacher-student pairs: ResNet101 $\to$ ResNet18 and ResNet101 $\to$ ResNet18. Specifically, we evaluate IC-KD using DeepLabV3-ResNet101 as the teacher and either DeepLabV3-ResNet18 or SPNet-ResNet18 as the student. To ensure fair comparisons, we adhere to the training protocol outlined in CIRKD~\cite{yang2022cross}, including image cropping to 512 x 512 pixels and employing random flipping and scaling in the range of [0.5, 2] for data augmentation. 

Given that semantic segmentation entails dense predictions, we acquire the positive and negative in-context features for each feature within one epoch. As presented in Tab.~\ref{segmentation}, our IC-KD outperforms existing KD methods, yielding superior segmentation performance across various student networks with similar or different architecture styles. Furthermore, IC-KD demonstrates significant improvements over CIRKD~\cite{yang2022cross}, achieving an average mIoU gain of 0.51 and 0.23 in these two tasks, respectively. These results underscore the efficacy of regularizing student training with in-context samples, facilitating effective knowledge transfer from teacher to student.


\section{Conclusion and Limitations}
\label{conculsion}

\textbf{Conclusion.} In this paper, we rethought `knowledge' in distillation from an in-context sample retrieval perspective and proposed a novel in-context knowledge distillation approach to regularize the student training with retrieving in-context samples. Our method introduces two regularization loss terms: positive in-context distillation and negative in-context distillation. Extensive experiments and ablation studies validate the effectiveness of our proposed approach. 

\textbf{Broader impacts.} IC-KD enhances compact the student performance through regularization with in-context samples, facilitating significant reductions in computational resources needed for inference and real-time data processing, without relying on cloud infrastructure. However, the KD process poses a risk of propagating biases and errors from the teacher model to the student model. Also, retrieving in-context samples from large-scale datasets may impose increased resource demands.

\textbf{Limitations and future work.} 
Despite its effectiveness and generality, IC-KD has two limitations. The primary limitation is the computational cost of applying retrieval-based learning to obtain in-context samples within the feature memory bank. Additionally, our approach rethinks `knowledge' in distillation based on logits, leaving the enhancement of student training through regularization in the feature space unexplored. Future work will investigate `knowledge' based on feature representation for knowledge distillation.

\bibliographystyle{IEEEtran}
\bibliography{IEEEabrv, egbib}

\vfill

\end{document}